\documentclass[
]{ceurart}

\usepackage{graphicx}
\usepackage{multirow}
\usepackage{comment}
\usepackage{amsmath}
\usepackage{amssymb}
\numberwithin{equation}{section}
\usepackage{algorithm}
\usepackage[noend]{algpseudocode}
\usepackage{float}
\usepackage{lettrine}
\usepackage{graphicx}
\usepackage{subcaption}
\usepackage{xspace}
\usepackage{makecell}

\begin{document}

\title{LAVA: Granular Neuron-Level Explainable AI for Alzheimer's Disease Assessment from Fundus Images}

\author[1]{Nooshin Yousefzadeh}[%
email=nooshinyousefzad@ufl.edu
]

\address[1]{Department of Computer \& Information Science \& Engineering, University of Florida, Gainesville, Florida, USA}

\author[2]{Charlie Tran}[%
email=charlietran@ufl.edu
]

\address[2]{Department of Electrical and Computer Engineering, University of Florida, Gainesville, Florida, USA}

\author[3]{Adolfo Ramirez-Zamora}[%
email=aramirezzamora@ufl.edu
]

\address[3]{Department of Neurology, University of Florida, Gainesville, Florida, USA}

\author[4]{Jinghua Chen}[%
email= jinghuachen@ufl.edu
]

\address[4]{Department of Ophthalmology, University of Florida, Gainesville, Florida, USA}

\author[2,5,6]{Ruogu Fang}[%
email=ruogu.fang@bme.ufl.edu
]
\cormark[1]

\address[5]{J. Crayton Pruitt Family Department of Biomedical Engineering, University of Florida, Gainesville, Florida, USA}
\address[6]{Center for Cognitive Aging and Memory, University of Florida, Gainesville, Florida, USA}

\author[1]{My T. Thai}[%
email=mythai@cise.ufl.edu
]
\cormark[1]

\cortext[1]{Corresponding author.}


\begin{abstract}
Alzheimer's Disease (AD) is a progressive neurodegenerative disease and the leading cause of dementia. Early diagnosis is critical for patients to benefit from potential intervention and treatment. The retina has been hypothesized as a diagnostic site for AD detection owing to its anatomical connection with the brain. Developed AI models for this purpose have yet to provide a rational explanation about the decision and neither to infer the stage of the disease's progression. Along this direction, we propose a novel model-agnostic explainable-AI framework, called Granu{\underline{la}}r Neuron-le{\underline{v}}el Expl{\underline{a}}iner (LAVA), an interpretation prototype that probes into intermediate layers of the Convolutional Neural Network (CNN) models to assess the AD continuum directly from the retinal imaging without longitudinal or clinical evaluation. This method is applied to validate the retinal vasculature as a biomarker and diagnostic modality for Alzheimer’s Disease (AD) evaluation. UK Biobank cognitive tests and vascular morphological features suggest LAVA shows strong promise and effectiveness in identifying AD stages across the progression continuum.

\end{abstract}



\maketitle

\lettrine{A}lzheimer's disease is the leading cause of dementia. The number of people aged 65 and older with AD in the United States is estimated to be around 6.5 million, which is expected to grow to 13.8 million by 2050 \cite{ADPROJECTION}.  AD is a progressive disease that can be broadly characterized into preclinical, prodromal mild cognitive impairment (MCI due to AD), mild AD, moderate AD, and severe AD based on the presence of clinical biomarkers and cognitive symptoms \cite{Aisen2017-om, ADCONVERSION}. Early screening and diagnosis of AD are essential to alter the disease trajectory. 

Pathological changes to the retina have been associated with early-stage neurodegenerative diseases \cite{shi2021retinal,koronyo2021retinal,  ong2014retinal}. Retinal screening presents a non-invasive, feasible, and economical solution to early AD diagnosis which has been hindered by the lack of consistent clinical symptoms and the absence of clinically accessible neuroimaging and biological markers \cite{redundancy}. Among the retinal features, weakening and alterations of the retinal vasculature as an AD biomarker have recently emerged \cite{Zhang2021-tv}. Clinical studies have focused on the time-consuming manual segmentation of the vasculature, propagating subjective error into the quantitative analysis. To counteract this problem, AI-based models have been introduced as more objective, repetitive, precise, and automated systems to aid the vasculature segmentation and the decision-making of ophthalmologists. Only few AI-based models have investigated AD through the retina \cite{wisely2022convolutional, MODULAR, Zhang2021-ac} and no work has yet studied the retinal biomarkers for AD across the disease spectrum. Furthermore, these AI-based models have been used as black-box models without a clear understanding of why the model made such predictions.

Recent advances in Explainable-AI (XAI) have shed interpretability into AI models. Notable explainers are feature attributions (e.g., saliency maps \cite{simonyan2019deep}, SHAP \cite{lundberg2017unified}, LIME, \cite{Ribeiro2016WhySI}, and integrated gradients \cite{sundararajan2017axiomatic}). In particular, these explainers are effective at the macro-level (e.g., input-wise or layer-wise) highlighting the features that are most effective in decision-making. However, these explainers lack information attained at the micro-level of artificial neurons which influences different mechanisms of decision-making. We look to invoke this ideology into the perspective of AD, that is, a medical XAI framework to identify sub-types and progression stages of the disease.

We propose our XAI framework called Granu{\underline{la}}r Neuron-le{\underline{v}}el Expl{\underline{a}}iner (LAVA) for explainable diagnosis and assessment of the AD continuum. The intuition behind this approach is that analyzing the behavior of neurons generates rich information reflecting not only the correlation between biomarkers but also the interaction among biomarkers, thanks to inductive learning of deep neural network architectures. We thereby introduce latent representations of raw pixels reflected in the activation behavior of neurons as a resource to discover and reveal hierarchical taxonomies of potential biomarkers. LAVA is a systematic approach that probes into intermediate layers of the CNN model, inspects and leverages the activation patterns of neurons as auxiliary information to improve model Explainability and Diagnostic power jointly. Subsequently, we show how this new source of information during the learning process is used to predict coarse-to-fine class in a downstream classification task where only coarse-level target labels are available; such discovered knowledge can be linked to the domain of knowledge to gain new insights from experts in the application domain.

\begin{figure}[!htbp]
\centering
    \includegraphics[scale=0.25]{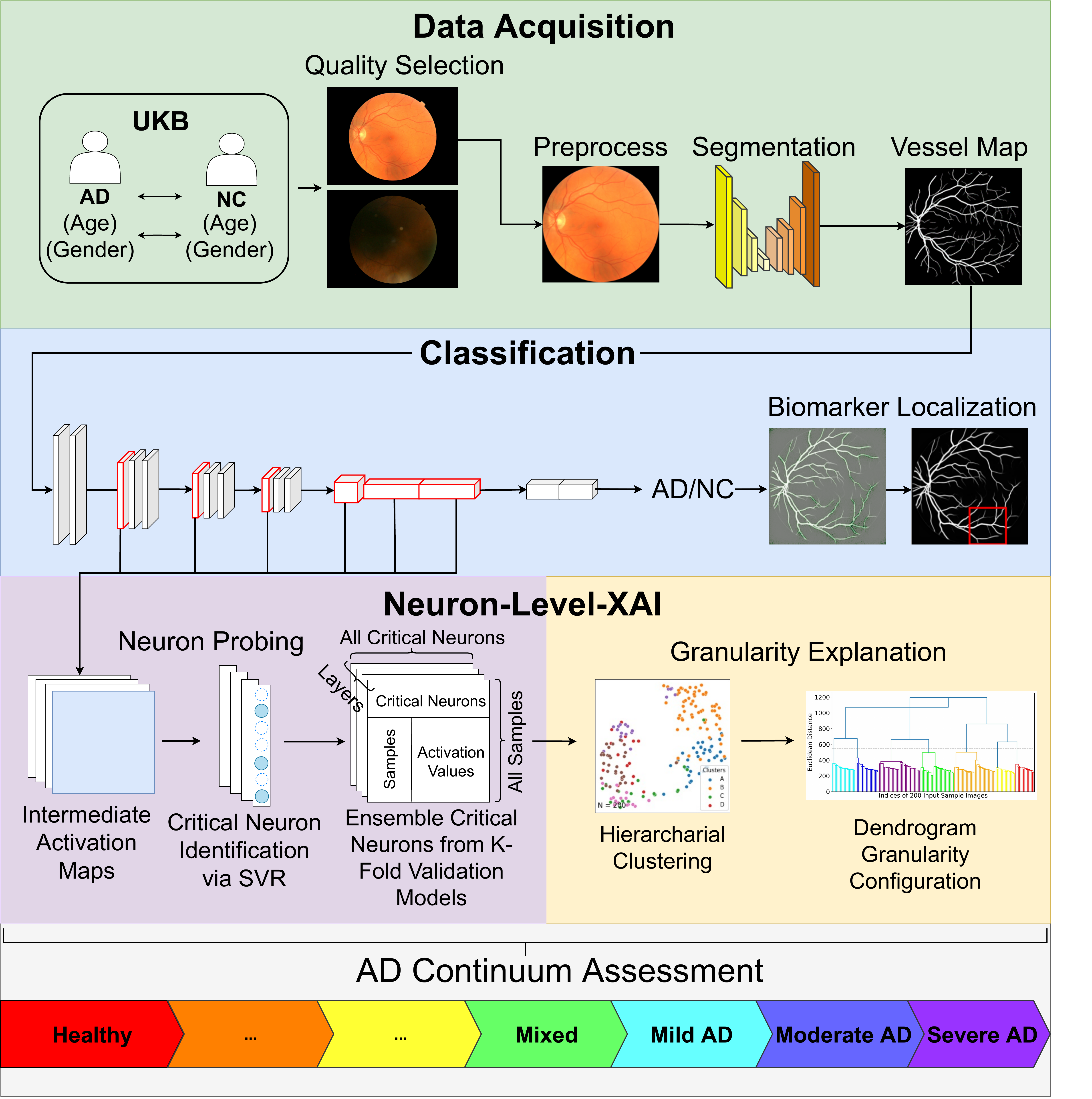}
    
    \caption{\textbf{Overall architecture of LAVA framework.} End-to-end learning process in LAVA framework is constituted by four main phases: (1) \textbf{Data Acquisition} where fundus images from the UK Biobank are collected along with quality selection and AutoMorph preprocessing to obtain retinal vasculature maps and morphological features, (2) \textbf{Classification} where a VGG-16 model is utilized for binary classification between AD and NC images supplemented with a feature attribution map, (3) \textbf{Neuron-level XAI} that consists of two modules of \textit{Neuron Probing} to identify and extract critical neurons across the VGG-16 network and \textit{Granularity Explanation} to identify sub-classes of labels hidden in data, and (4) \textbf{AD Continuum Assessment} where the diagnostic result of LAVA for an individual subject is summarized.} 
    \label{fig:diagram}
\end{figure}

There are two core modules so-called \textit{Neuron Probing} and \textit{Granularity Explanation} that constitute the LAVA architecture, as shown in Fig. \ref{fig:diagram}. The former identifies critical neurons and inspects their activation patterns. The latter clusters input sample images into distinctive groups emulated by activation of critical neurons as independent random variables. LAVA is input size invariant, model-agnostic in the sense that it can adapt to a broad class of CNN models\footnote{The activation of neurons are extracted during the test phase, hence CNN models that do not contain any dropout layers in their architectures are preferable in this framework in order to avoid randomized and non-reproducible results.}, and adjustable to the granularity level of data in the application domain. 

In this article, we present the development of a novel XAI framework, Granular Neuron-level Explainer (LAVA) to evaluate fundus images as a diagnostic modality of AD continuum assessment. We verify the effectiveness of LAVA through consistency checks using clinical measures of cognitive function and vascular integrity in the UK Biobank \cite{sudlow2015uk}. We employ feature attribution and pixel reconstruction methods to highlight regions of interest in the diagnosis of AD. The proposed framework supports the automation of an XAI diagnostic system which may be used for clinical intervention and advance the field's mechanistic understanding of AD.

\section*{Results}\label{sec1} 
\noindent\textbf{Study design and participants.} LAVA is developed to assess AD classification and infer the disease continuum  utilizing fundus images acquired from the UK Biobank \cite{sudlow2015uk}. The UK Biobank contains nearly 170,000 fundus images from over 500,000 participants. Quality control is performed to exclude fundus images with artifacts and clarity issues using a pretrained CNN module on the EyePacs-Q dataset. AD subjects with other additional sub-types of dementia (e.g., frontotemporal dementia (FTD) are excluded. We identify a total of 100 images from 61 unique AD subjects. To avoid potential confounding factors, we construct our binary-labeled dataset by matching each AD image with 80 unique age and gender-matched normal controls (NC) leading to a total number of 200 fundus images. The AutoMorph deep learning pipeline \cite{zhou2022automorph} is used for preprocessing, vessel segmentation, and morphological vascular feature quantification. 

\noindent\textbf{Training and inference.} A VGG-16 \cite{simonyan2014very} binary classifier model is trained and evaluated under five-fold stratified cross-validation setting on the segmented vessel maps. This procedure is repeated with five repetitions with an optimal 5-fold accuracy of 75\% and average accuracy of 71.4\% (SD = 0.03). The best-performing cross-validation model is utilized for post-hoc analysis.

\noindent\textbf{Neuron Probing.} We probe into intermediate layers of the network at the neuron level to assess the AD continuum (see Fig. \ref{fig:diagram}). In our setting, we chose Max-Pooling layers and the first two fully connected layers of the VGG-16 for critical neuron selection. Although our approach supports critical neuron extraction from early layers, we find our LAVA framework works effectively using a combination of Max-Pooling Layers. Owing to the Maximum Likelihood Estimation (MLE) algorithm to approximate the joint Mutual Information (MI) objective in the selection of critical neurons, the LAVA framework is reproducible, model agnostic, and input size invariant. We use Epsilon-support vector regression ($\epsilon$-SVR) \cite{Frey2007ClusteringBP} with a linear kernel as a core algorithm to estimate the contribution coefficient of every single neuron at selected layers and wrap the output by Recursive Feature Elimination (RFE) to collectively realize our MLE-based critical feature selection objective.

We set two hyperparameters for the number of selected critical neurons at each layer to be 20 and the number of neurons pruned at each iteration to be 1000. This MLE-based feature selection procedure repeats to ensemble five sets of selected neurons at each layer by each of five cross-validation models into approximately 700 critical neurons, 100 from each layer, concatenated (with repetition) across the network. Supplementary Fig. \ref{fig: S1} shows the Jaccard similarity index computed to compare overlapping between sets of neurons selected by five cross-validation models. Higher similarities suggest similar activation behavior at the certain layers which can be interpreted as similar Region Of Interest (ROIs) used for the feature extraction. 

\noindent\textbf{Granularity Explanation.} Using the results obtained in neuron probing, we can distill the activation values of critical neurons across the network over all input samples as a new dataset that will be used for our knowledge discovery. Under a semi-supervised setting, LAVA employs the  Adjacency-constrained Hierarchical Agglomerative Clustering (HAC) algorithm \cite{Ambroise2019AdjacencyconstrainedHC} where an early constructed $k$ nearest-neighbor graph ($k-$NNG) imposes connectivity constraints in the form of a 97.5\% ($k=5, N=200$) sparse connectivity matrix of shape $N\times N$ that links each input sample to its five nearest neighbors. This algorithm first creates a distance matrix for sample instances using the Euclidean metric and then reduces a chunk of distances to the $k$-nearest neighbors for each sample where the array of distances for that sample is partitioned by the element index $k-1$ in the stable sorted order. $k$ is a hyperparameter chosen based on experience and the number of target labels; $k=3$ and $k=5$ are common choices in the LAVA framework. The results show this approach is highly effective in using fundus biomarkers to identify latent sub-classes of predicted label interpreted as AD continuum.   

\begin{figure}[!htbp]
\centering
\includegraphics[scale=0.4]{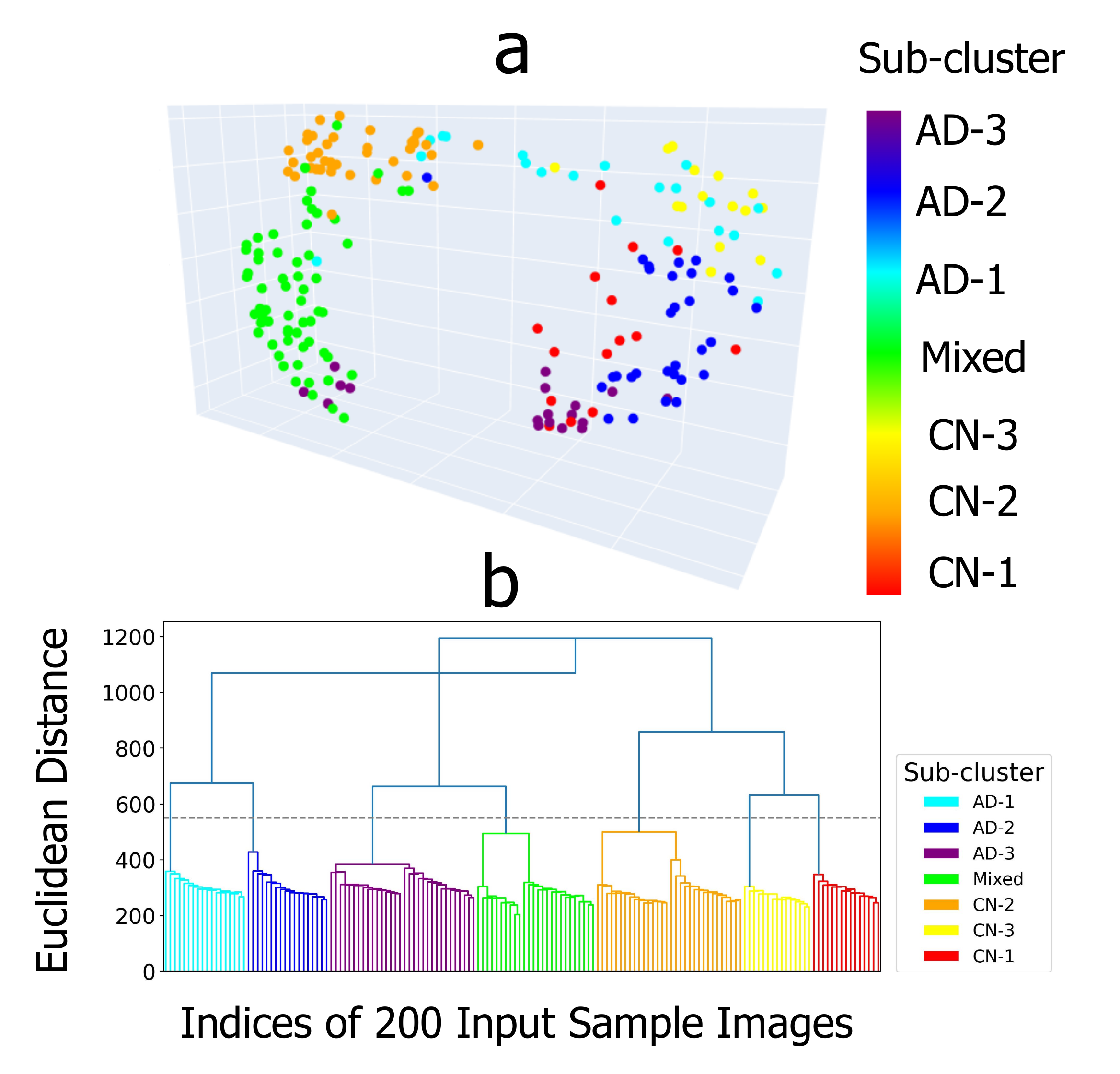}
\caption{\textbf{Neuron-level probing results.} \textbf{(a)} 3D visualization of UMAP embedding for high-dimensional data sample points. Samples with similar embeddings (close to each other or similar in the behaviour of critical neurons) have similar sub-cluster label predicted by HAC algorithm that effectively reveals the clustering structure within data. \textbf{(b)} Dendrogram of agglomerative connectivity constraint clustering with Ward's Linkage represents the similarity relationship among sub-clusters of AD subjects in terms of the behaviour of critical neurons. Imaginary horizontal line traversing dendrogram determines the correspondent detail level of latent sub-clusters that characterizes subjects within the continuum of disease.}
\label{fig: probing}
\end{figure}

Fig. \ref{fig: probing}\textbf{(a)} (Supplementary Fig. \ref{fig: S2}) visualizes UMAP (Uniform Manifold Approximation and Projection for Dimensionality Reduction) \cite{McInnes2018UMAPUM} embedding of input fundus images in terms of their activation of critical neurons projected in three dimensions. Unlike t-SNE, UMAP does not completely preserve density of data and thus provides a more effective preprocessing tool for our clustering. In this study, we make use of UMAP visualization and the dendrogram diagram (Fig. \ref{fig: probing}\textbf{(b)} and Supplementary Fig. \ref{fig: S3}) for two purposes: (1) Initial evaluation of the hardness of the clustering task, and (2) Decision on the appropriate number of clusters. The number of clusters is a hyperparameter in LAVA framework normally chosen based on the granularity level of the data and the nature of the problem under study, which is set to 7 in this experiment. We used intrinsic metrics e.g., Calinski-Harabasz (CH) index (also known as the Variance Ratio Criterion) \cite{calinski1974dendrite} and Adjusted Mutual Information (AMI) \cite{Nguyen2009InformationTM} to choose the appropriate clustering method by comparing their performances. The result of this clustering is summarized in Supplementary Table \ref{table: table1} including 3 purely AD groups, 3 purely NC groups and 1 Mixed group of coarsely AD or NC labeled subjects.

\begin{figure}[!htbp]
\centering
\includegraphics[scale=0.2]{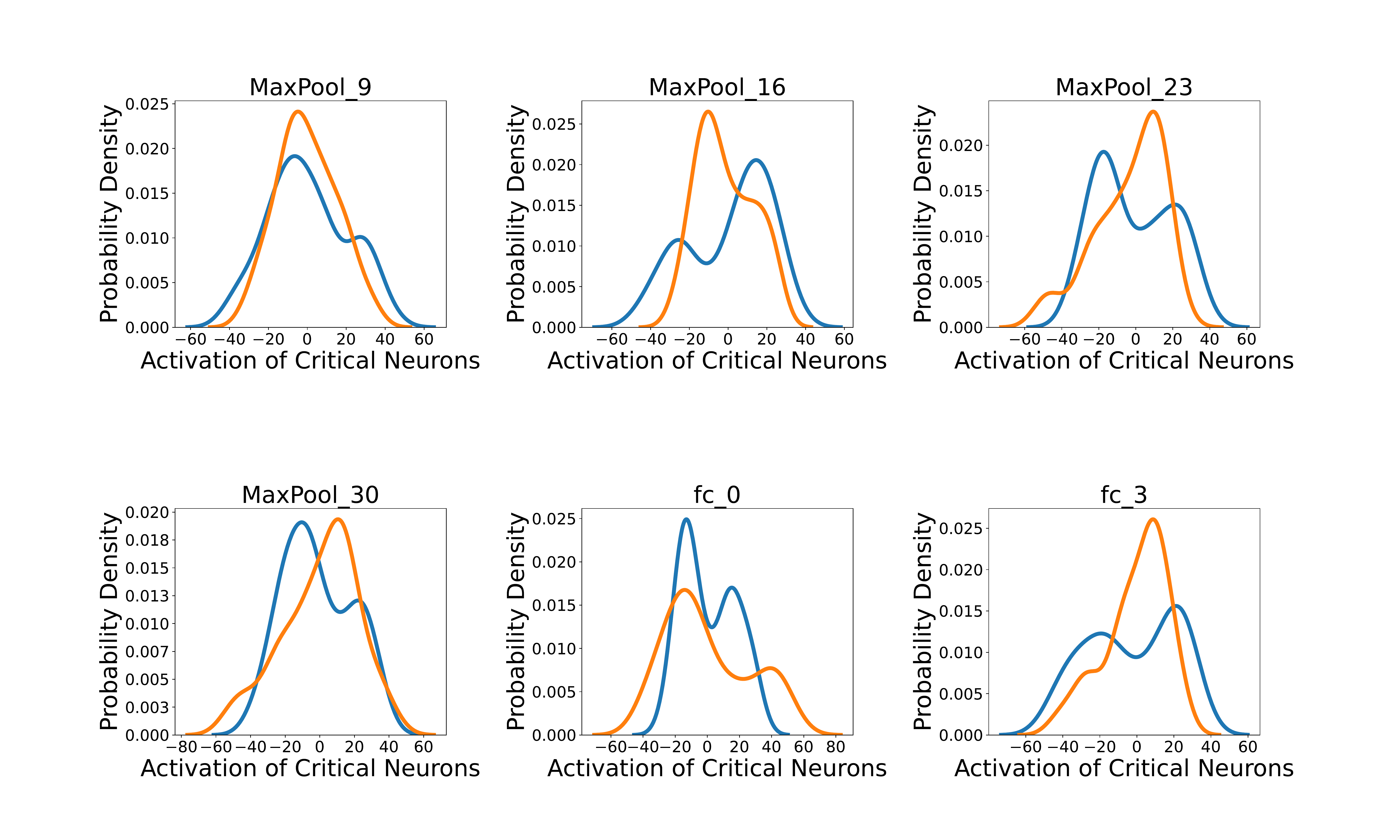}
\caption{\textbf{Exploring activation pattern of critical neurons.} KDE applied on top of two-dimensional t-SNE embedding of critical neurons' activation values unveils multiple activation patterns for the same set of critical neurons at each layer. It suggests evaluated fundus images belong to different sub-cluster of patients within AD continuum associated with AD or NC target class (blue and orange curves) of the disease.}
\label{fig: t-sne}
\end{figure}

We further analyzed the behavior of critical neurons of the trained network independent from the input data. First, we use t-SNE  \cite{van2008visualizing} to project high-dimensional space of critical neurons' activation values at a certain layer down to two dimensions. Second, we apply Kernel Density Estimation (KDE) \cite{chen2017tutorial} method on top of t-SNE to estimate the probability density curve associated with each dimension of t-SNE embedding as shown in Fig. \ref{fig: t-sne} and Supplementary Fig. \ref{fig: S4}. Blue and orange curves at each layer can be interpreted as the distinctive behavior of the model in the prediction of coarse class labels (AD/NC), while the presence of multiple peaks at each curve reveals a mixture of multiple probability distributions corresponding to the different mechanisms of prediction or different activation patterns used in the prediction of each class of label. Our intuition is that each peak can potentially correspond to one distinctive sub-cluster of examined samples. This observation is an analogy to previous UMAP and dendrogram visualization of latent clustering structure within 200 input samples suggested by activation values of critical neurons in the network.

The choice of hierarchical clustering over other semi-supervised clustering methods e.g., KMeans \cite{Arthur2007kmeansTA}, Mean Shift Clustering \cite{Comaniciu2002MeanSA}, Affinity Propagation \cite{Frey2007ClusteringBP}, etc. is made based on the behavior of critical neurons and how well each clustering algorithm can scale on our dataset. We use various statistical methods e.g., Variance Ratio Criterion, Adjusted Mutual Information, Rand index, V-measure, homogeneity score, and completeness score to evaluate and compare the performance of different clustering algorithms. We observe medical assessments reported on UK Biobank cognitive tests \cite{UKBCOG} efficiently scale over a hierarchy and not a flat set of clusters. The primary results encourage our further investigation into finding appropriate clustering algorithm in order to gain more insights on learning the connection between AD-related biomarkers in eye fundus images and activation pattern of critical neurons in the network.

\noindent \textbf{Continuum assessment.} Next, we showcase our LAVA-based hierarchical clustering is reflective of the AD continuum. Seeing that the UKB lacks detailed assessment of activities of daily living, cognitive profile or functional scores (e.g., the clinical dementia rating and the mini-mental state examination) and neither brain imaging data in our cohort, we use cognitive test measures from the UKB as proxy measures of cognitive ability \cite{spindola2011prospective}. These tests  include two-levels of memory from the UKB, the pairs matching and prospective memory, and an intellectual problem-solving measure, the fluid intelligence. We note that the clusters are extracted from retinal vasculature images, and thus, we hypothesize that image-level features should coincide with our continuum. Naturally, such image-level features live in an abstract space. To resolve this issue, we evaluate quantifiable morphological features, specifically, the fractal dimension and vessel density, that are representative of the image-level features and relate these to the cognitive ability of a subject.

\begin{figure}[!htbp]
\centering
\includegraphics[scale=0.1]{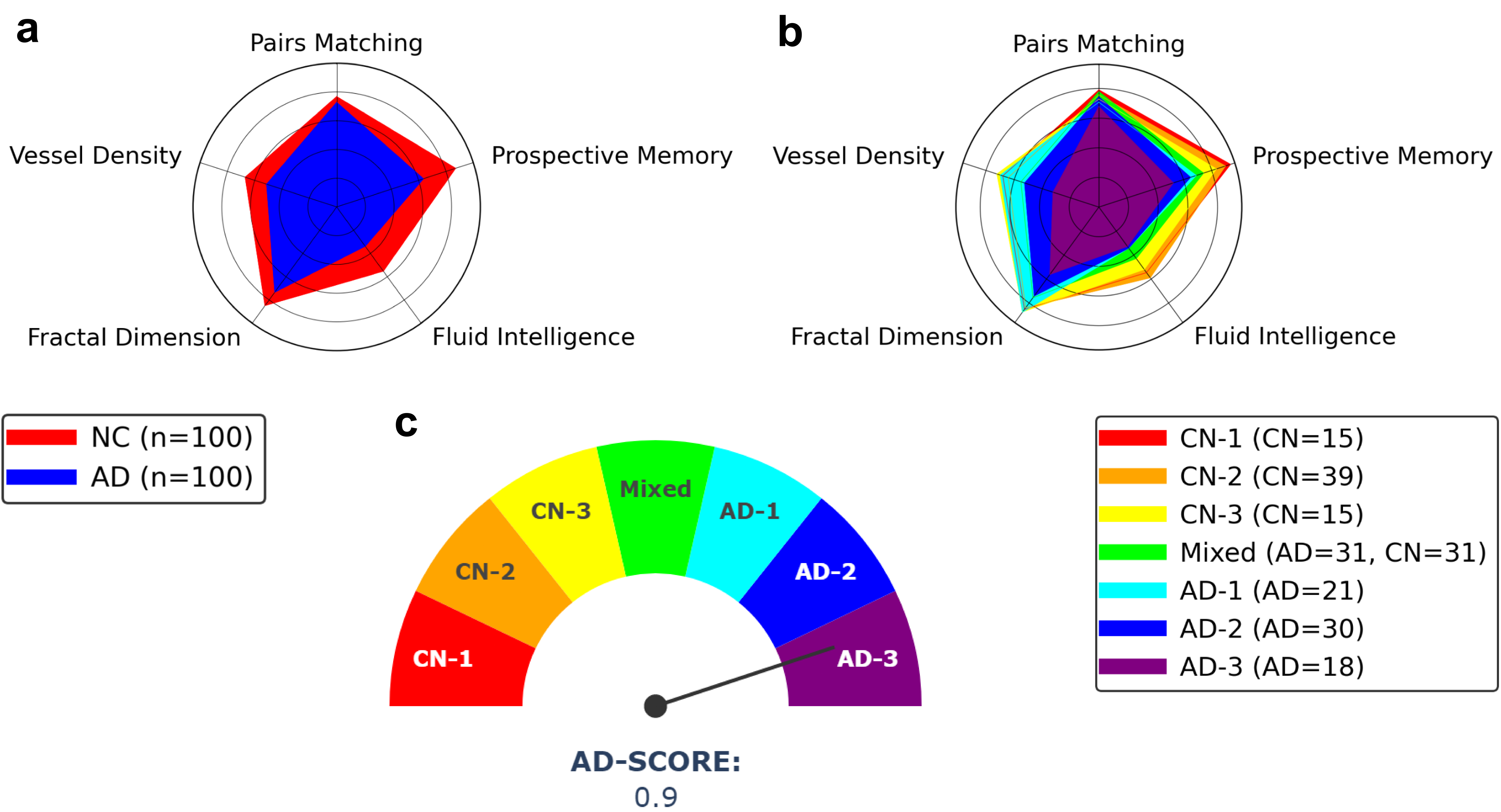}
\caption{\textbf{LAVA evaluation with clinical measures.} The cognitive and vascular features are normalized onto [0,1] for scalable weighting. \textbf{(a)} The UK Biobank cognitive test measures and vascular features between AD and NC groups. \textbf{(b)} The cognitive and vascular feature comparison in the continuum identified by the LAVA framework. \textbf{(c)} The AD-score defined by averaging the normalized cognitive features.  Each sub-group block is not drawn to scale.}
\label{fig: cluster_analysis}
\end{figure}

We employ our analysis at the group-level (AD/NC) and the sub-group level. First, we verify that the cognitive measures are significantly different across groups, as shown in Supplementary Fig. \ref{fig: S5}.  Next, as each metric is on a different scale, all of the scores are normalized on [0,1] for comparison. A normalized comparison across groups is illustrated in Fig. \ref{fig: cluster_analysis} \textbf{(a)} through a visual radar plot. We demonstrate that such metrics form an increasing sequence of measures across clusters, supporting the idea that such latent clusters are indicative of the AD continuum. From this observation, we term our sub-groups in order ranging from the healthiest states of cognitive normal (CN) to the severity of AD [CN-1, CN-2, CN-3, Mixed, AD-1, AD-2, AD-3]. Notably, the Mixed Group contains a sub-cohort of AD and NC subjects suggesting similarities of AD subjects and potentially at risk NC subjects. Furthermore, the reduction in morphological vascular features coincides with decline in cognitive ability, thus supporting the retinal vasculature as indicative of the AD continuum, as shown in Fig. \ref{fig: cluster_analysis} \textbf{(b)}. Last, taking from the observation the sequence of our clusters, we look to assign a simplistic AD score of the continuum as illustrated in the gauge plot, Fig. \ref{fig: cluster_analysis} \textbf{(c)}. To accomplish this, we average together the normalized cognitive metrics (pairs matching, prospective memory, and fluid intelligence). In this manner, the healthiest subject has a score of 0 and a severe subject has an upper bound score of 1. Therefore, for any new subject, we may apply our LAVA framework and assign a subject's vessel map a position in the AD continuum as a manner for assigning their risk and potential clinical intervention.

\begin{figure}[!htbp]
\centering
\includegraphics[width=1\columnwidth]{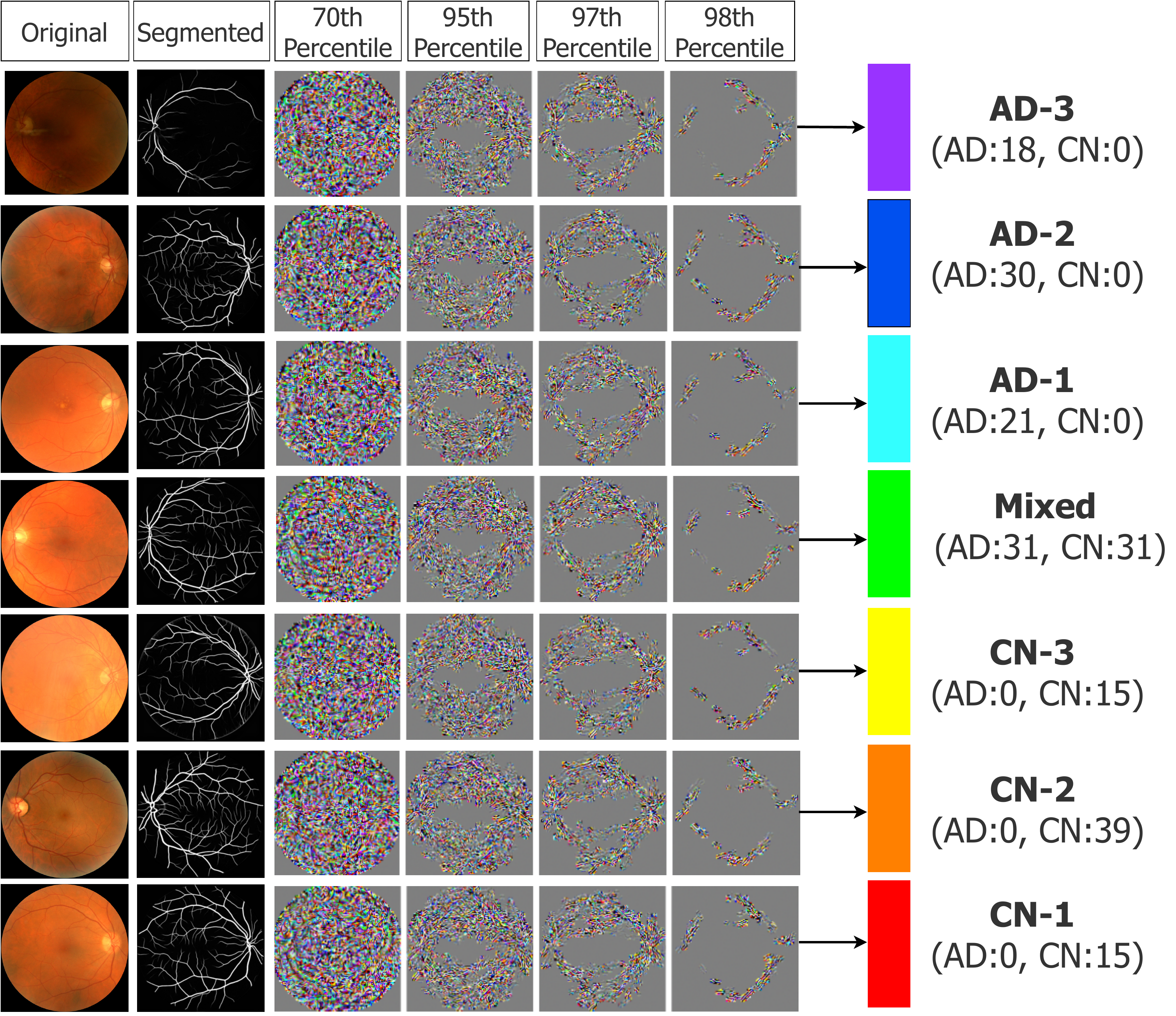}
\caption{\textbf{Gaining insights into determinative biomarkers.} The original CNN network is masked to include only a percentile of important neurons identified by LAVA, while Integrated Gradients computes the gradient of the model's prediction output with respect to its input features. The reconstructed pixels reflects the layout of biomarkers highly associated with the prediction of AD continuum progression. As the subnetwork becomes sparser for the most critical neurons, the reconstructed pixels reveal the most critical biomarkers effective in the diagnosis of each sub-cluster of the prediction.}
\label{fig:fig10}
\end{figure}

\noindent \textbf{Visual model interpretation for clinical evaluation.} We investigate the learning process by use of the guided backpropagation method, wherein we mask crucial input features and examine how the essential ROIs effective in the discovery of the AD continuum develop. With some modifications to the pruning objective, we use the method introduced in \cite{Khakzar2021NeuralRI} to reconstruct critical fragments effective in the prediction of each latent sub-class using a sparse pathways limited to some percentile of critical neurons identified and scored in Neuron-level XAI phase. In this technique, the Integrated Gradients method \cite{sundararajan2017axiomatic} is combined with Lucent objective \cite{lucent}, and as shown in Fig. \ref{fig:fig10}, biomarkers can be decoded at different levels of criticality to highlight the most determinant regions in the AD continuum prediction prioritized from the most specific to the most general.

\begin{figure}[!htbp]
\centering
\includegraphics[scale=0.5]{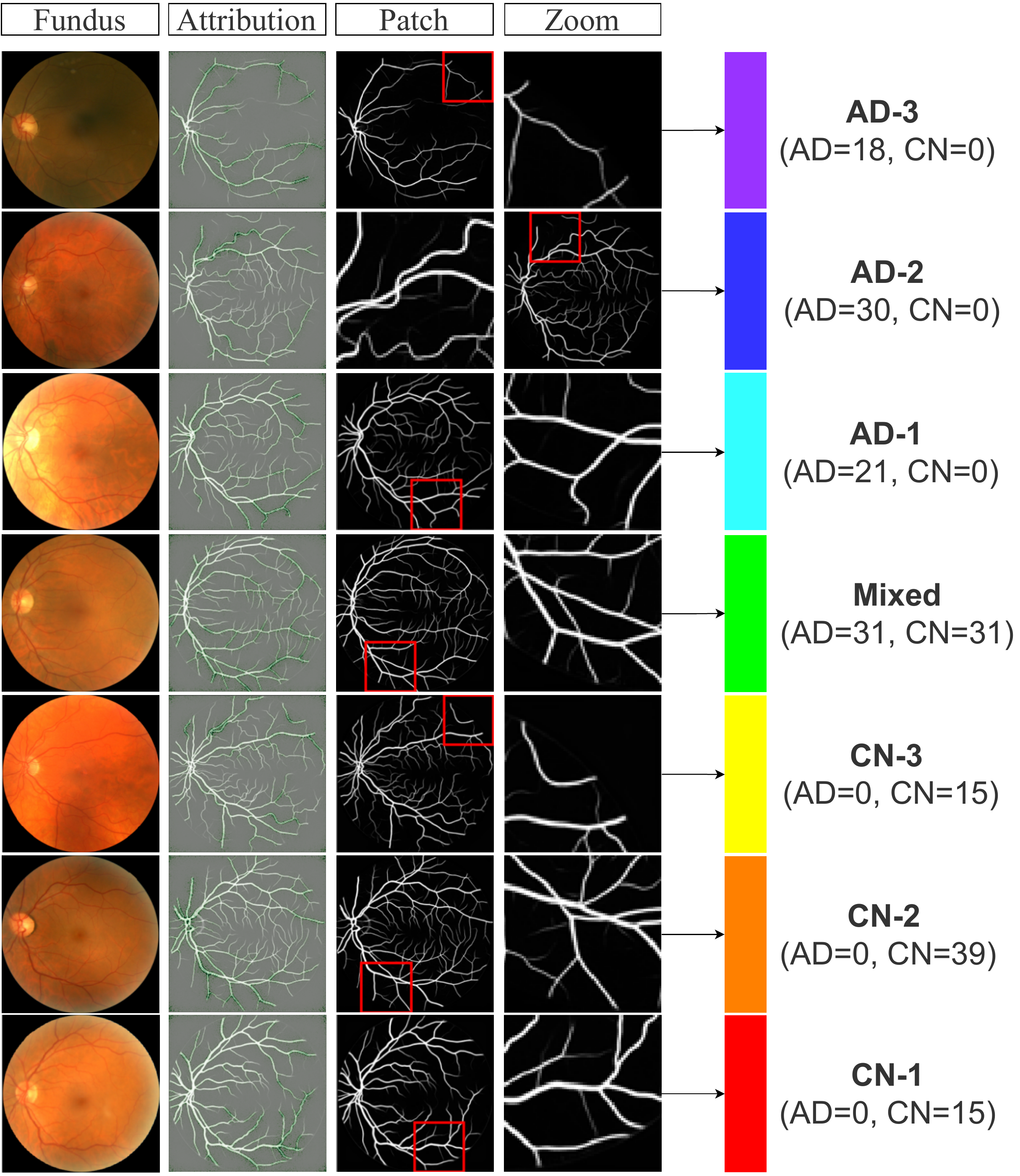}
\caption{\textbf{Visualization of input features identified relevant to the prediction}. The guided back-propagation method is employed to identify the region of interest. A sliding window is used to identify a crop of the image with the highest feature attribution. The reference fundus image (column 1) and the zoomed-crop (column 4) are shown for diagnostic visualiation  that may help to explain the vascular biomarkers across the predicted continuum of AD progression.}
\label{fig:gbp}
\end{figure}

Furthermore, we apply the prior technique in conjunction (and in comparison with) traditional attribution maps achieved by guided backpropagation \cite{selvaraju2017grad}, to develop an effective method for searching relevant biomarkers at different scales. The guided backpropagation is repeated using the Noise Tunnel Algorithm \cite{DBLP:journals/corr/SmilkovTKVW17} averaged 10 times for robustness of relevance. To  mimic a clinician's diagnostic decision-making, we zoom into a $70 \times 70$ crop of the image of highest feature attribution (see Fig. \ref{fig:gbp}). Nevertheless, while the GBP visualization reveals where to place attention for clinical observers, a true understanding of visual biomarkers remains unclear and requires future research collaborated with domain experts in neuro-ophthalmology. For this reason, we hope that a combination of visual model interpretation and quantifiable morphological features can be used together for informed judgement.


\noindent \textbf{Sanity Check of the Explanation.} We evaluate the faithfulness of the LAVA in providing a true explanation of the model's behaviors. We feed LAVA with a VGG-16 binary classifier where the parameters are randomized and examine how much the set of critical neurons in this model differs from that of original model. We observe a significant change in the set of critical neurons identified by LAVA at each layer of the model after the weights of the network are replaced by random weights (Jaccard similarity index computed as at fc-3 layer is 0.008 and zero at every other layers). This suggests LAVA truly extracts neurons critical to the output of the model i.e., extracted neurons are correctly explaining the behavior of the network.

\section*{Discussion}\label{sec2}

We develop Granular Neuron-Level Explainer (LAVA), a novel explainability framework for AD classification from fundus imaging. Specifically, we equip a traditional VGG-16 CNN with a five-fold cross validation binary classification accuracy of 75\% with a neuron-level XAI framework to support the retinal vasculature as an efficient AD screening modality. Our explanations are generated through a two-phase-procedure: (1) neuron probing and (2) granularity explanation. Notably, the utilization of a neuron-level-XAI model is valid, as the contribution of neurons is a better representation of the human imperceptible input features than the contribution of the raw input image pixels themselves \cite{Khakzar2021NeuralRI}. The reason behind this argument is that the hidden (latent) variables constructed during the learning process by convolutional deep neural networks play a significant but underrated role in characterizing and fully describing the undergoing phenomenon in medical image processing.

Few prior studies using AI models have been approached using retinal fundus images \cite{wisely2022convolutional, MODULAR, Zhang2021-ac}. However, these models do not consider the different stages of AD progression and thus do not offer a comprehensive evaluation of the risk severity. One of the primary contributions of this work is the projected inference of binary class labels (AD/NC) into latent sub-classes, which we claim to be indicative of the AD continuum. We support this argument through a number of approaches through comparisons of cognitive tests, morphological vascular features, and several visualizations modules. The value of our findings is applicable in many biomedical applications to enhance the interpretation of classifier models, allowing enhanced diagnostic judgement and understanding of the underlying biological phenomena. 

To demonstrate our claim of continuum assessment and allow various levels of enhanced interpretability, we compare the differences across cognitive level features (e.g., the pairs matching test, prospective memory, and fluid intelligence), morphological feature (fractal dimension and vessel density), and visualization models. All designated cognitive and vascular measures are demonstrated to be reduced in the AD group compared to the normal controls at statistical significance ($p < 0.01$). We extend these differences from a group-level to a latent-sub-group level via visual gauge and radar plots to demonstrate a sequence of clusters, ranging from healthiest to strongest severity of AD. In particular, we are capable of arranging the latent-sub-groups identified by our LAVA framework into a seven-level continuum, and design a simplistic manner for assigning an \textit{AD-score} to a subject. Guided backpropagation maps and critical neuron reconstruction techniques are used to determine diagnostic biomarkers, regions of interest, and differences amongst subgroups as deemed important by the model.

Although our study presents the ability to assess the AD continuum, the framework carries limitations. First, the amount of data is lacking, with only 100 images from 61 AD participants total, which hinders our model training and generalization to the real-world setting. Furthermore, we do not consider the effects of other confounding factors (demographics, genetics, etc.) or similarities in retinal biomarkers amongst other neurodegenerative diseases. While our work supports the reduction in cognitive performance and vessel structures \cite{ong2014retinal, ALZFRACTAL} in the retina, large individual variations between subjects could hinder the retina as a diagnostic site and need further validation. We also acknowledge the limitations of current cognitive variables and limited data assessing cognitive domains affected early and late in the disease like episodic memory and language. On the other hand, our XAI work supports clinical studies associating retinal degeneration to be linked to Alzheimer's Disease, rather than normal aging, as well as connections with cerebral small vascular disease.

Overall, our study demonstrates an explainable and systematic framework to map subjects into the progression continuum of  Alzheimer’s Disease using retinal vasculature from fundus images. Our method is effective in enhancing biological and diagnostic understanding, and automating healthcare streamlining and preclinical screening. This study will be helpful in examining how retinal pathology is connected to cognitive impairment neurodegeneration, with not only applications to AD, but other types of dementia and neurological/retinal diseases.

\section*{Methods}\label{sec3}
LAVA is a systematic method that leverages neuron-level explanation as auxiliary information during the learning process to predict coarse-to-fine class in a downstream classification task where only coarse-level target labels are available. In this section, the details of our proposed XAI framework are provided. 

\noindent \textbf{Neuron Probing.}
In the first phase of LAVA framework, we look to find a subset of critical neurons at each layer of the CNN model containing the most information concerning the prediction of class labels. 

Let consider any CNNs classification model $\Phi$ with a sequential structure consisting of $L$ layers, where each layer $l$ has $K_{l}$ neurons and $l=\{1,..,L\}$. Once any input sample $x \in \mathbb{R}^n$ is fed into the model $\Phi$ through the forwarding function $y=f(x)$ where $y \in \mathbb{R}^m$ is a logit, the activation of neurons at layer $l$ denoted as $Z_l$ is a random variable and also a function of the input $ Z_{l}=f_{l}(x)$ where $f_l : \mathbb{R}^n\rightarrow \mathbb{R}^{k_l}$. The forwarding structure of the neural networks suggests the activation of neurons at each layer depends only on the activation of the neurons at the previous layer i.e., $Z_l  {\!\perp\!\!\!\perp}  Z_i \vert  Z_{l-1}, \forall i= 0,...,l-2 $, where ${\!\perp\!\!\!\perp}$ denotes the independent relationship.

Our goal is to find a subset of critical neurons at each layer $l$ containing the most information on the prediction of interest. Recently, the notion of criticality in neuron-level extraction and the objective of critical neurons identification subsequently is formulated with joint mutual-information (MI) function \cite{vu2022neucept} from probability and information theories \cite{10.5555/1146355} to measure the mutual dependence between two variables.

Let $(Z_l,Z_{l+1})$ be two discrete random variables over the space $\mathcal{Z}_l \times {\mathcal{Z}_{l+1}}$ to indicate activation of neurons for a pair of adjacent layers in CNN model. If $P_{(Z_l,Z_{l+1})}$ denotes the joint distribution and $P_{Z_l}$ and $P_{Z_{l+1}}$ denote marginals, then the amount of information shared between those two adjacent layers can be measured by an MI objective that searches for the set of critical neurons at each layer on the set of critical neurons solved in the next layer \cite{vu2022neucept}.

\begin{equation} \label{eq:1} \tag{1}
MI(Z_l;Z_{l+1}) = \sum_{Z_l\in\mathcal{Z}_l }    \sum_{Z_{l+1} \in \mathcal{Z}_{l+1} } P_{(Z_l,Z_{l+1})}\log \frac{P_{(Z_l,Z_{l+1})}}{P_{Z_l}P_{Z_{l+1}}}
\end{equation}

Thus, a sequence of MI objectives, starting from the last layer, can be optimized at any layer with respect to its preceding layer to identify the most critical neurons from each layer $M_{l}$ through the network. This sparse sub-network of critical neurons conveys the most important information all the way from input to output of the model. 

\begin{equation} \label{eq:2} \tag{2}
M_l=\operatorname*{argmax}_{M_l \subseteq  K_{l}} MI(Z_l^{M_l} ;Z_{l+1}^{M_{l+1}})
\end{equation}

Directly solving Equation (\ref{eq:2}) at each pair of adjacent layers in this sequential optimization formulation is in NP-hard \cite{Davies1994NPCompletenessOS}, because as proved in \cite{vu2022neucept}, MIN-FEATURES \cite{davies1994np} problem can be reduced to this problem in polynomial time. On the other hand, the state of Markov chain of $L$ layers ($Z_0\rightarrow Z_1\rightarrow...\rightarrow Z_L$) \cite{pearl2014confounding} suggests $Z_l^{M_l}$ can determine $Y$, and consequently, $ M_l$ that contains $M_l(Y)$. As a consequence, to overcome the the curse of dimensionality, an approximation solution can solve MI objective at each pair of a layer with the output $(Z_l^{M_l} ;Y )$ instead of solving that at each pair of adjacent layers $(Z_l^{M_l} ;Z_{l+1}^ {M_{l+1}})$ as follows: 

\begin{equation} \label{eq:3} \tag{3}
M_l \approx \operatorname*{argmax}_{M_l \subseteq  K_{l}} MI(Z_l^{M_l} ;Y)
\end{equation}
The entropic (informational) correlation between a feature and class label in high-dimensional scheme is a useful statistic measurement for feature selection. As the mutual information enlarges, the feature becomes more significant and distinguishable. Let $z \in {Z_l}$ denote a feature (single neuron at layer $l$) and $y \in Y$ denote a class of label, then the mutual information between them can be defined as follows:

\begin{equation} \label{eq:4} \tag{4}
MI(Z_l;Y)=\sum_{j=1}^M \sum_{i=1}^N p(y_j,z_i)\log_2 \frac{p(y_j,z_i)}{p(y_j)p(z_i)}
\end{equation}

\noindent where $n=\{1,...,N\}$ and $m=\{0,...,M\}$ are the number of different values for $z$ and $y$ respectively, $p(z_i)$ and $p(y_j)$ are marginals, and $p(z_i,y_j)$ is the joint probability.

NeuCEPT \cite{vu2022neucept} uses Model-X Knockoffs as a statistical tool with false discovery rate control to approximate Markov Blanket \cite{koller2009probabilistic} as the smallest subset of neurons at each layer maximizing the MI; however, it imposes some limitations in our application: (1) The subtle differences among fundus images result in low variance in distribution of neurons' activation which makes it difficult for any neurons to be selected. (2) Selection of critical neurons from large-sized intermediate layers of our network is difficult due to the complexity of matrix inversion operation involved. In order to overcome aforementioned limitations, MI can be alternatively approximated using density estimates \cite{Suzuki2008ApproximatingMI} based on the Kernel Density Estimator (KDE) \cite{chen2017tutorial} and thus Model-X Knockoffs can be replaced with any Maximum Likelihood Estimation (MLE)-based feature selection technique as an estimation of MI. 

We adopt the same method used in \cite{Guyon2004GeneSF} for gene selection from expansive patterns of gene expression data in genetic diagnosis (or drug discovery) in LAVA to capture a very small and compact (non-redundant) multiset\footnote{An ensemble bag with repeated elements (multiplicity is allowed.)} of the most critical neurons at each selected layer through the network while evaluating the binary target labels (AD/NC) by cross-validation models in the different subsets of input images. This approach combines Epsilon Support Vector Regression with Recursive Feature Elimination algorithm ($\epsilon$-SVR+RFE) to satisfy the MLE objective in selection of the critical neurons across all $L$ layers each of $K_l$-dimensionality. Following the same objective of Joint Mutual Information (MI) function, $\epsilon$-SVR \cite{Platt2007ProbabilisticOF} maximizes likelihood estimation to identify critical
neurons with respect to the output of the model. More specifically, this feature scoring method constructs a coefficient vector with a logit link function and a regularized maximum likelihood score. Thus, as shown in \cite{lin2012support}, this compact feature selection technique employed in LAVA assures that the neuron whose MI is larger is more likely to be selected as critical. In this technique, RFE is a wrapper-type statistical method that uses $\epsilon$-SVR algorithm in the core. It eliminates the least important features iteratively until the desired number of features is reached. 

Epsilon-supported Support Vector Regression ($\epsilon$-SVR) attributes coefficients of contribution to each neuron under acceptable maximum error $\epsilon$ (epsilon). First the activation value of $M$ critical neurons at a selected layer for $N$ sample inputs  $\{Z\}^{N \times M}$ maps in feature space $U=\phi(Z$), and then a hyperplane is constructed using a kernel function $f(Z,W)=W^TZ+b$ that minimize its deviation from training data by minimizing L2 norm of the coefficient vector $\vert \vert W \vert \vert$. The setting of hyperparameters for $\epsilon$-SVR includes a kernel parameter (e.g, linear, sigmoid, radial basis function, and polynomial) and a regularization parameter. The latter is used to make a tradeoff between the complexity of the model and the accuracy of the training. The objective function of $\epsilon$-SVR is as follows:

\begin{equation} \label{eq:5} \tag{5}
\min \frac{1}{2} {\vert\vert W \vert\vert }^2 +C \sum_{i=1} ^{n} \vert \xi_i \vert
\end{equation}

\noindent constrained by $\vert y_i - \sum _{j=1}^M w_j(z_{i,j})\vert \leq \epsilon+ \vert \xi_i \vert$. 

Here $w_j$ is the coefficient of the support vector in the decision function assigned to the $j$-th critical neuron and $\epsilon$ denotes a margin for absolute distance value between actual and predicted values in the training loss function for which no penalty is associated. Penalties can be regularized by $C$ as a measure of tolerance for the output of the $i$-th input sample to fall outside $\xi_i$ deviation from true output variable $y_i$ and still is acceptable within error margin $\epsilon$.
The implementation of $\epsilon$-SVR is from LIBSVM library \cite{Chang2011LIBSVMAL}.

The results obtained by $\epsilon$-SVR+RFE feature selection technique at the Neuron Probing phase of LAVA framework can be adjusted to desired detail level that characterizes subpopulations within the target continnum. The pseudocode of this algorithm is provided in Supplementary Algorithm \ref{alg:alg1}.

\noindent \textbf{Granularity Explanation.}
In the second phase of the LAVA framework, we search to answer this question{\em ``To what extent similarity between input samples in terms of the pattern of activation is consistent with that in terms of true labels in the multi-granularity deep local structure of target domain?''} 
To discern the division of AD subjects, our  choice of agglomerative connectivity constraint clustering with ward's linkage (also known as Minimum Variance) is rational to the intrinsic granularity of diagnostic biomarkers associated with the continnum of progressive nature of Alzheimer's disease. Our hypothesis is that hierarchical representation levels hidden in the input image dataset can interpret biomedical features associated with progressive disease in this study. Although such artifacted dynamics biomarkers mapped to different levels of abstraction representation made by a convolutional deep neural model might not be visually descriptive and explicitly perceptible by humans.

We employ the Hierarchical Agglomerative Clustering (HAC) method with connectivity constraints ($k$-NNG graph) \cite{Ambroise2019AdjacencyconstrainedHC} to partition entire eye fundus image samples into subgroups of data based on similar activation patterns of critical neurons during evaluation of the image by the CNN model. The description and pseudocode of this algorithm is provided in Supplementary Algorithm \ref{alg:alg2} and Supplementary Method \ref{sec:apx4}.

The appropriate number of clusters can be visualized and heuristically determined on a hierarchical ward's tree of the clustering learning process so-called dendrogram. The height of rectangles fit between different levels of hierarchy in dendrogram represents the distinctiveness among clusters at that specific level. The number of vertical lines cut by an imaginary horizontal line traversing dendrogram determines the correspondent number of clusters as a configurable parameter in HAC learning algorithm. 

\noindent \textbf{Study Population and Baseline Characteristics } 
This study is conducted on the UK Biobank. At the time of acquisition (Jun 2019) of our UKB basket, there are a total of 1,005 AD subjects from approximately 500,000 total subjects. In particular, we investigate the incidence of AD, which is defined as subjects who are diagnosed with Alzheimer's Disease after the baseline visit, in contrast to prevalent AD, which consists of subjects with a record of AD diagnosis before the baseline visit (as according to the UK public health records, ICD9 and ICD10 codes). From the 1005 AD subjects in the UKB, there are 111 AD subjects with fundus images. Next, we manually select an overall number of 100 images from 61 unique AD subjects images upon the following criteria: (1) the incidence of AD, (2) sufficient visibility of the retinal vasculature in terms of the level of artifacts and clarity of the image, and (3) no record of other neurodegenerative diseases, excluding subjects with mixed dementia and/or forms of Parkinsonism. To prevent external bias in our analysis, we perform age and gender matching for each AD subject with a normal control (NC). We note that a normal control subject is taken with the understanding of no current label of dementia, regardless of whether a subject may be at risk, or develop dementia in the future. We identify 100 images from a total of 80 unique NC subjects, wherein an additional NC subject is substituted when the image quality for certain fundus image pairs for a matched AD subject is insufficient.

Supplementary Table \ref{table: study_population} showcases the summary statistics of the study population including demographics (age, gender, and ethnicity), ophthalmic features, and covariates. The ophthalmic features include eye problems (e.g., glaucoma, cataratcs, etc.) and visual acuity (LogMAR).  The  covariates include townsend indices, obesity-diabetes status, smoking status, alcohol status, and history of stroke. Obesity-diabetes is defined as a BMI greater than 30 or diagnosis of diabetes. All baseline characteristics were selected on the basis of explored risk factors in AD-studies \cite{Gong2021, Lumsden2020}.

\noindent \textbf{Data-Preprocessing.} A manual image quality selection is employed to ensure that each fundus image has sufficient retinal vascular visibility. We employ the AutoMorph pipeline for image pre-processing. In particular, the image undergoes thresh-holding, morphological image operations, and cropping to effectively remove the background of the fundus images. The images are passed into a Segan \cite{xue2018segan} network for vessel segmentation pretrained on a collection of labeled retinal vasculature datasets. The details of pre-training can be found in \cite{zhou2022automorph}. During the model training and evaluation of our classifiers, the vessel maps are resized to $224 \times 224$ in accordance with ImageNet. The vascular morphological features are computed using the original image size.

\noindent \textbf{Data Partitioning, Tuning, and Evaluation. } 
In the binary classification task, we apply nested stratified five-fold cross validation. To avoid potential bias, each fold is split such that eyes from the same subject are contained in the same fold and each fold is equivalent in number (n = 20 images). To maximize the limited data, we tune the hyper-parameters during training with a four-fold cross validation loop and re-train the model over all training data with the best hyper-parameters. Our experiments suggest optimization over a small learning rate grid of [1e-4, 1e-5] and a maximal number of 50 epochs is sufficient. We use a cross-entropy loss, Adam optimizer \cite{kingma2014adam}, and data augmentations (flipping and rotations) for fine-tuning an ImageNet pre-trained VGG-16 classifier \cite{simonyan2014very}. The model is then re-trained over the optimal hyper-parameters using all of the training data and tested on the outer cross-validation fold.

\noindent \textbf{Vascular Morphological Feature Measurements.} The AutoMorph  pipeline is used to extract the vessel density and fractal dimension \cite{Falconer2010-zo} from the retinal vasculature. These measures are chosen on the basis of hypothesized mechanisms concerning the reduction of vessel structures (e.g., small vessel disease) and structural complexity \cite{VESSELDENSITY, CVD, ong2014retinal}. The vessel density is defined as the proportion of vessel pixels to the number of pixels in the image. The fractal dimension is defined here as the Minkowski-Bouligand dimension, also known as the box-counting dimension. Let $X$ denote a (square) image, that is, the input vessel map. The Minkowski-Bouligand dimension is thus defined as follows:
\begin{equation} \label{eq:6} \tag{6}
\textrm{FD}_{box} (X) = \lim_{\epsilon \rightarrow 0} \frac{\log (N/\varepsilon) }{\log (1/\varepsilon) }
\end{equation}

\noindent Discretely, $N$ is the input size of the image where $\epsilon$ is taken such that the window size of the box is reduced by a factor of $2$ until the window attains a box of chosen size, $16 \times 16$.

\noindent \textbf{Cognitive Tests.} 
The UKB administers several cognitive tests for a subset of the UKB cohort \cite{UKBCOG}.
The Pairs Matching Test (Field 399; Number of Incorrect Matches) contains a three card and six card variant. We select the six card variant for analysis for simplicity, larger variance, as well as being used in other studies \cite{Lyall2016}. Moreover, we extract the prospective memory (Field 20018: Prospective Memory Result), and fluid intelligence (Field 20016: Fluid Intelligence Scores). Overall, these tests are chosen on the basis of natural associations of reduction in AD subjects (symptomatic of the loss of memory and problem-solving) which have  been investigated in prior clinical studies \cite{Spindola2011-zl, Raz2008-gp}. For the few subjects who do not have cognitive test measures, their values are interpolated using the average over their diagnostic class.



\section*{Acknowledgments} 
This material is based upon work supported by the National Science Foundation under Grant No. (NSF 2123809).

\section*{Data Availability Statement} This research has been conducted using the UK Biobank Resource under application number 48388. The datasets are available to researchers through an open application via \url{https://www.ukbiobank.ac.uk/register-apply/.}

\newpage
\bibliography{bibliography}

\newpage
\begin{center}
\section*{Supplementary Information}
\title{LAVA: Granular Neuron-Level Explainable
AI for Alzheimer’s Disease Assessment from
Fundus Images}
\end{center}

\appendix
\section{Supplementary Methods}
\noindent \textbf{Adjacency-constrained Hierarchical Agglomerative Clustering.}\label{sec:apx4} Hierarchical clustering can be generated either top-down called \textit{divisive clustering} similar to $k$-means (where a data set is divided into more number of smaller clusters gradually) or bottom-up called \textit{agglomerative clustering} (where initially every data point is considered as an individual cluster and then gradually merged into less number of bigger clusters). Divisive clustering can be linear in the number of clusters if the number of top levels is fixed, despite that the number of clusters in LAVA formulation is not pre-defined and depends on the application and the granularity nature of the data structure. We use Hierarchical Agglomerative Clustering (HAC) with ward's linkage. The time complexity of naive agglomerative clustering is $O(n^3)$ and can be reduced to $O(n^2 logn)$ when priority queue data structure is used and can be reduced to $O(n^2)$ with some further optimization.
In HAC algorithm, the between-cluster agglomerative distance can be recursively computed, while aggregated distance between clusters can be updated without need to compute all the pair of objects contained in the clusters. In this setting, we use ward's linkage to update aggregated distance between clusters. This approach attempts to merge two clusters for which the change in total variation is minimized. The total variation of a clustering result is the sum of squared-error $ESS(C)$ (so-called \textit{inertia of cluster C} \cite{Ambroise2019AdjacencyconstrainedHC}) between every object and the centroid of the cluster containing that object. Thus, Ward’s linkage criterion $\delta$ can be formulated as follows when two clusters $C$ and $C'$ are merged where $\Bar{C}$ is the the mean vector (centroid) of the clusters.

\begin{equation}
\delta(C, C') = ESS(C\cup C')-ESS(C)-ESS(C')
\end{equation}

\begin{equation}
ESS(C) := \frac{1}{\vert C \vert}\sum_{i\in C} \vert\vert x_i-\Bar{C}\vert\vert^2
\end{equation}

Suppose we have two clusters $C$ and $C'$ that are merged into a new cluster $C^*$, and let $C''$ be any other cluster. Let the size of cluster $C$ , $C'$ , $C''$ be $n_c$, $n_c'$, $n_c''$ correspondingly. Algorithm updates distance $D(C^*, C'')$ from $D(C, C'')$ and $D(C', C'')$ through a systematic pairwise
distance $D(x_i, x_j )$ for every $i\not=j$ is given as follows.

\begin{equation} \label{eq:14}
\begin{split}
\tiny
&D(C^*,C'')=\frac{n_c+n_c''}{n_c+n_c'+n_c''}D(C,C'')+\\
&\frac{n_c'+n_c''}{n_c+n_c'+n_c''}D(C', C'')- \frac{n_c''}{n_c+n_c'+n_c''}D(C, C')
\end{split}
\end{equation}
 
In our experiment, we use the entire set of input images for which the array of features (critical neurons' activation) serves as raw training data so-called activation dataset in the constrained version of clustering algorithm in semi-supervised setting. 

Given $k$ different parameterization of a classifier model $\Phi$ through nested $k$-fold cross-validation learning paradigm, with $L'$ selected layers where  $l'=\{1,..,L'\}$, and $N=\{1,...,N\}$ total number of input samples $X=\{x_1, ...,x_n\}$. let ${\{Z_{l'}^k\}}_{i=1}^N$ denotes the activation of critical neurons at selected layer $l'$ of $k$-th model for $i$-th input instance. To aggregate activation values, first we stack activation values across all cross-validating models $\{Z_{l'}\}_{i=1}^N$. Second we stack them across all selected layers $\{Z\}_{i=1}^N$ (where $i$ is an index of input sample instance) to construct a two-dimensional array of the activation values of critical neurons across entire  networks over all input samples. 

Let $Y=\{y_1, ...y_n\}$ denotes the array of ground-truth labels for all input images. We construct the connectivity graph $h=\{0,1\}^{N\times N}$ out of the $k$-nearest neighbor graph (K-NNG) as constraints in semi-supervised learning algorithm. In this graph, if the distance between two nodes $p$ and $q$ is among the $k^{th}$ smallest distance from node $p$ to any other nodes, $p$ and $q$ are connected. In this setting, standard Euclidean metric measures the difference of ground truth labels assigned to each sample point. The output of this algorithm is a sparse CSR-format connectivity matrix A of shape $N \times N$ where only $k \times N$ number of entries (self-included) are one and the rests are zero. This algorithm reduces a chunk of distances to the $k$-nearest neighbors where elements are partitioned by element index $k-1$ in the stable sorted array of distances for each sample instance.

Connectivity constraints make the clustering algorithm performs differently in the constrained version of HAC in two aspects:

\begin{enumerate}
	\item After each step of merging, a graph ${h^{(p)}}$ will be created (recursively) to record the connectivity constraints between clusters at any iteration $p$ where current clusters are treated as nodes in the graph.
	\item Two clusters can be merged only if they are connected according to the connectivity constraint graph at the current iteration ${h^{(p)}}$.
\end{enumerate}
The pseudocode of this clustering method is provided in Supplementary Algorithm \ref{alg:alg2}

\section{Supplementary Algorithms}

\begin{algorithm}[H]%
\scriptsize
\caption{LAVA - Neuron-level Probing}\label{alg:alg1}
\begin{algorithmic}[1]
\State{\textbf{Input:} \\A binary vector $\hat{Y}=(\hat{y}_1,..\hat{y}_n)$ of predicted labels for $n$ input samples $X=(x_1,..x_n)$.\\A set of all neurons at $L$ layers denoted as $\{{S}_{l}\}_{l=1}^{L}$  and their activation values for all input samples, denoted as $\{{Z}_{l}\}_{l=1}^{L}$.\\
A positive integer $P$ number of critical neurons to extract from selected layer.}\\
A Kernel type.\\
An equal or greater than 1 integer for regularization parameter $C$.\\
\State { \textbf{Output:} \\Set of critical neurons at each layer ${\{\hat{S}_{l}\}}_{l=1}^{L}$ and their activation values ${\{Z'_l\}}_{l=1}^K$.}\\

\For {$l=1$ to $L$}\Comment{at each selected layer}
\State $scores \Leftarrow$ coefficient of contribution of neurons to the output of the model $\hat{Y}$ estimated by $\epsilon$-SVR on $\{{Z}_{l}\}_{l=1}^{L}$.
\State${\hat{S}}_l \Leftarrow$ recursively eliminate the least important neurons based on the $scores$ until $P$ is reached by RFE.
\State	$Z'_l \Leftarrow$ filter activation values for only critical neurons at each layer.
\EndFor
\State \textbf{return} $\hat{S}$ and $Z'$.
\end{algorithmic}
\end{algorithm}

\begin{algorithm}[H]%
\scriptsize
\caption{LAVA - Granularity Explanation}\label{alg:alg2}
\begin{algorithmic}[1]  
\State{\textbf{Input:}\\ An array $Z'$ of activation values of critical neurons at $L$ layers for all samples (obtained from Algorithm \ref{alg:alg1})\\
A positive integer $K$ number of nearest neighbors.\\
A binary vector $Y=(y_1,..y_n)$ of target labels for $n$ input samples $X=(x_1,..x_n)$.\\
Ward's linkage criterion $\delta$.\\
A pairwise distance metric $\zeta$.\\
A positive integer $R$ number of clusters.\\}
\State {\textbf{Output :} \\A ward tree representation $U$ of input samples.\\
A vector of cluster labels $W$ for input samples.}\\
  
\State  \# Linkage matrix  construction
\State	\hspace{3em}  $h={\{0,1\}}^{n \times n} \Leftarrow$ construct sparse connectivity matrix with $Y$ and $K$
\State  \# Semi-supervised learning 
\State	\hspace{3em} Construct distance matrix $M$ on $Z'$ with $\zeta$.
\State	\hspace{3em} $(C^0)=(C_i^0)_{1 \leq i \leq N}$ with $C_i^0 =\{x_i\}$\Comment{initializing clusters}
\State	 \hspace{3em} Initialize $h^{(0)}$ graph on clusters $C_i^0$ as vertices.
\State 	\hspace{3em} \textbf{for} $i=1$ to $n-1$ \\
\hspace{4em} \textbf{if} $(C_{u}^{(i-1)}, C_{u+1}^{(i-1)}) \in h^{(i-1)}$  \Comment{find best merging candidate}
\State 	\hspace{6em} $d_t=argmin_{d\in\{1,...,N-i\}} \delta(C_u^{i-1}, C_{u+1}^{i-1})$
\State 	\hspace{6em} Update graph $h^{(i)}$  on new clusters.
\State 	\hspace{6em} Update matrix $M$ by removing row and column of merged clusters.
\State 	\hspace{6em} \textbf{for} $d=1$ to $n-i-1$ \Comment{update $C^i$ with $C^{i-1}$}
\State 	\hspace{9em} \textbf{if} $d<d_i$ then $C_u^i =C_u^{i-1}$
\State 	\hspace{9em} \textbf{else if} $d=d_i$ then $C_u^i =C_u^{i-1}\cup C_{u+1}^{i-1} $
\State 	\hspace{9em} \textbf{else if} $d>d_i$ then $C_u^i =C_{u+1}^{i-1}$
\State 	\hspace{9em} \textbf{end if}
\State 	\hspace{6em} \textbf{end For}
\State 	\hspace{3em} \textbf{end for}
\State \hspace{3em} $g \Leftarrow $ constructed hierarchical ward tree
\State 	\hspace{3em} Fit $g$ on $Z'$ for R clusters.
\State 	\hspace{3em} $U \Leftarrow $ transformation of $Z'$ by ward tree $g$
\State 	\hspace{3em} $W \Leftarrow $ predict sub-class labels on $Z'$ by ward tree $g$
\State \textbf{return} $U$ and $W$.
    
\end{algorithmic}
\end{algorithm}




\section{Supplementary Figures}

\begin{figure}[!htbp]
\centering
\includegraphics[width=1\linewidth]{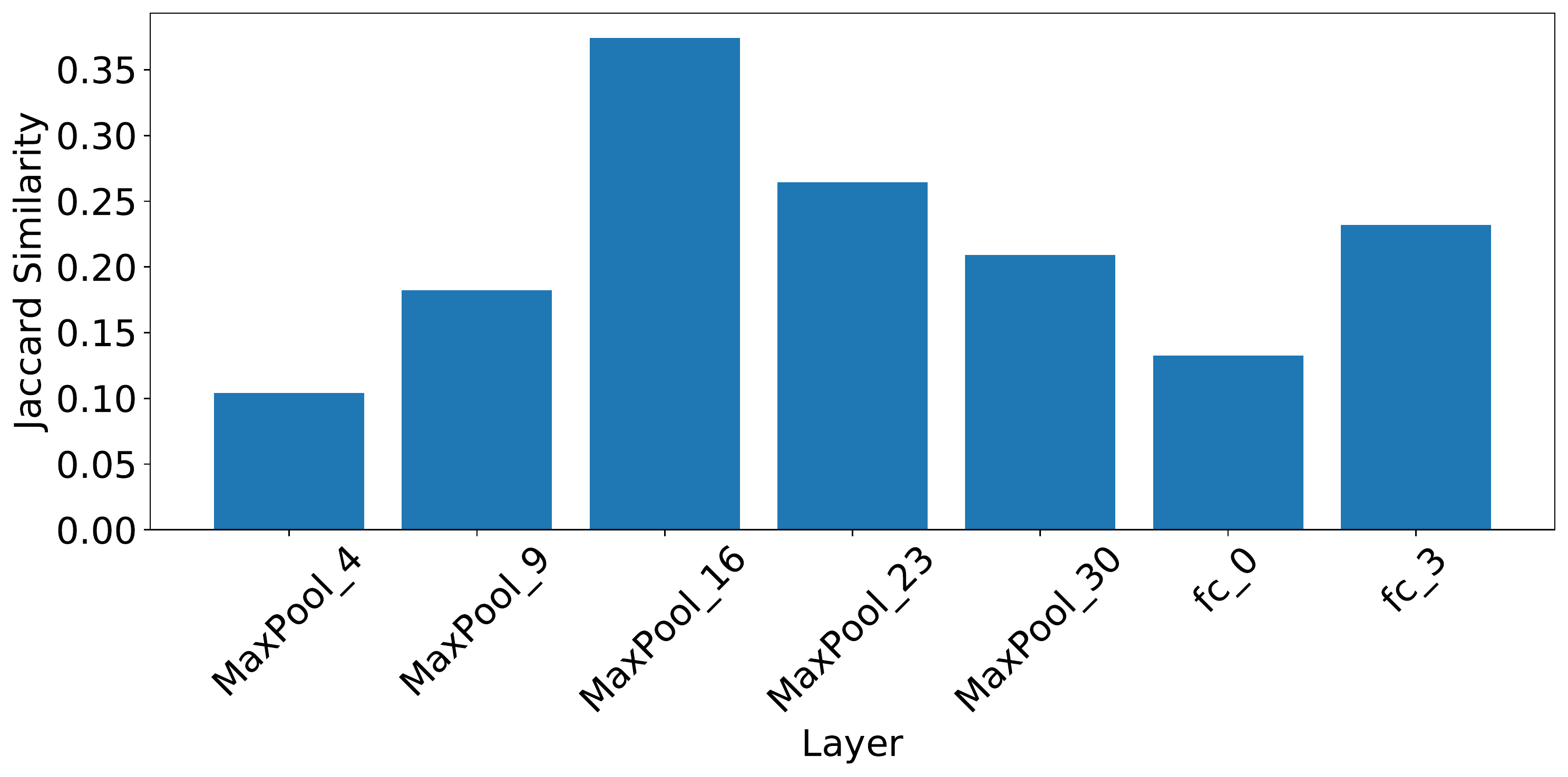}
\caption{\textbf{Critical neurons identification.} Overlapping between sets of critical neurons at different layers of the network identified repetitively by different parameterizations of the model obtained from K-fold cross-validation is measured by Jaccard similarity.}
\label{fig: S1}
\end{figure}

\begin{figure}[!htbp]
\centering
\includegraphics[width=1\linewidth]{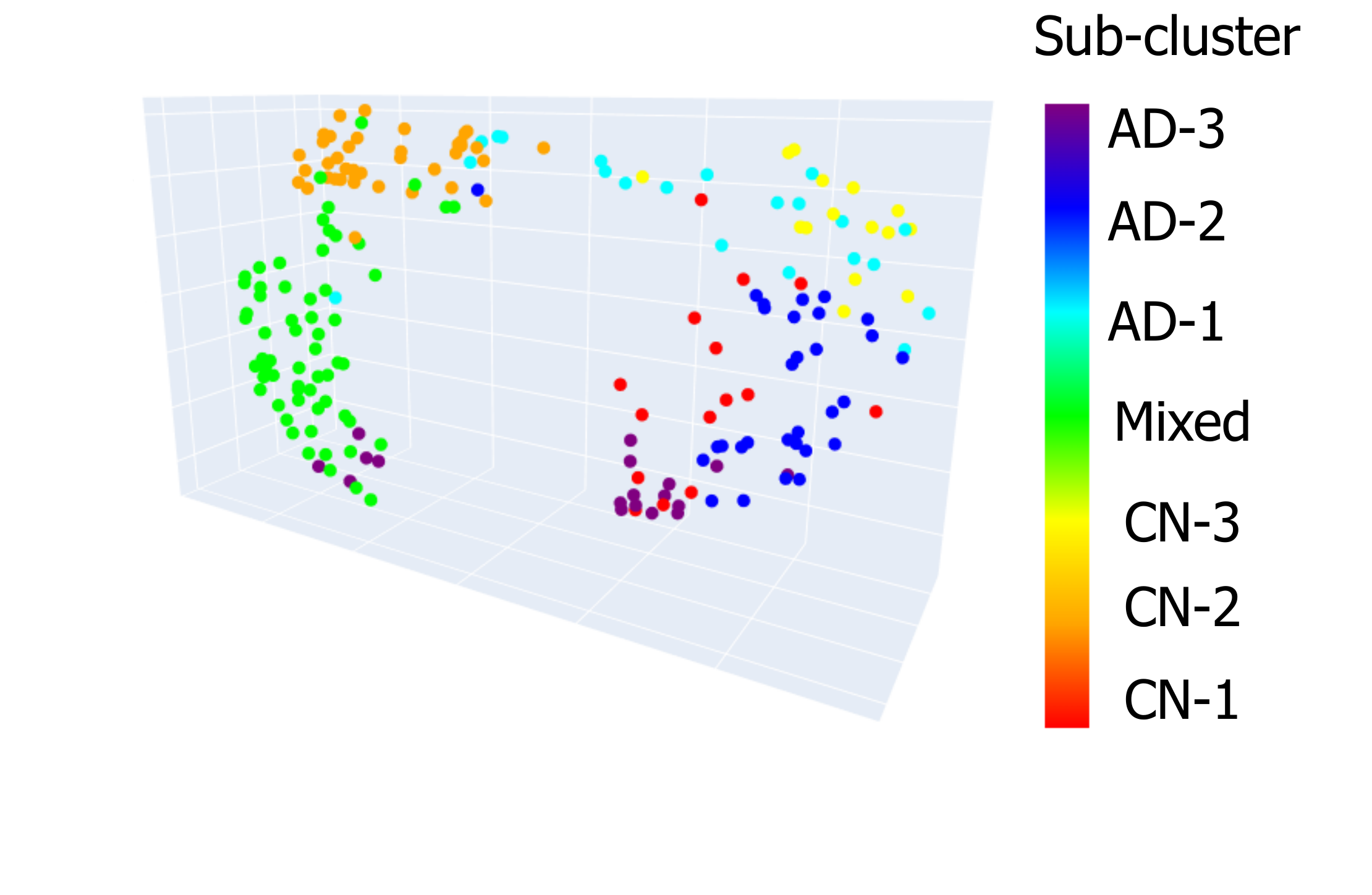}
\caption{\textbf{UMAP embedding of input image samples.} UMAP embedding reveals potential sub-clusters within each target class of label learned from eye fundus images at LAVA's neuron-level probing phase.}
\label{fig: S2}
\end{figure}

\begin{figure}[!htbp]
\centering
\includegraphics[width=1\linewidth]{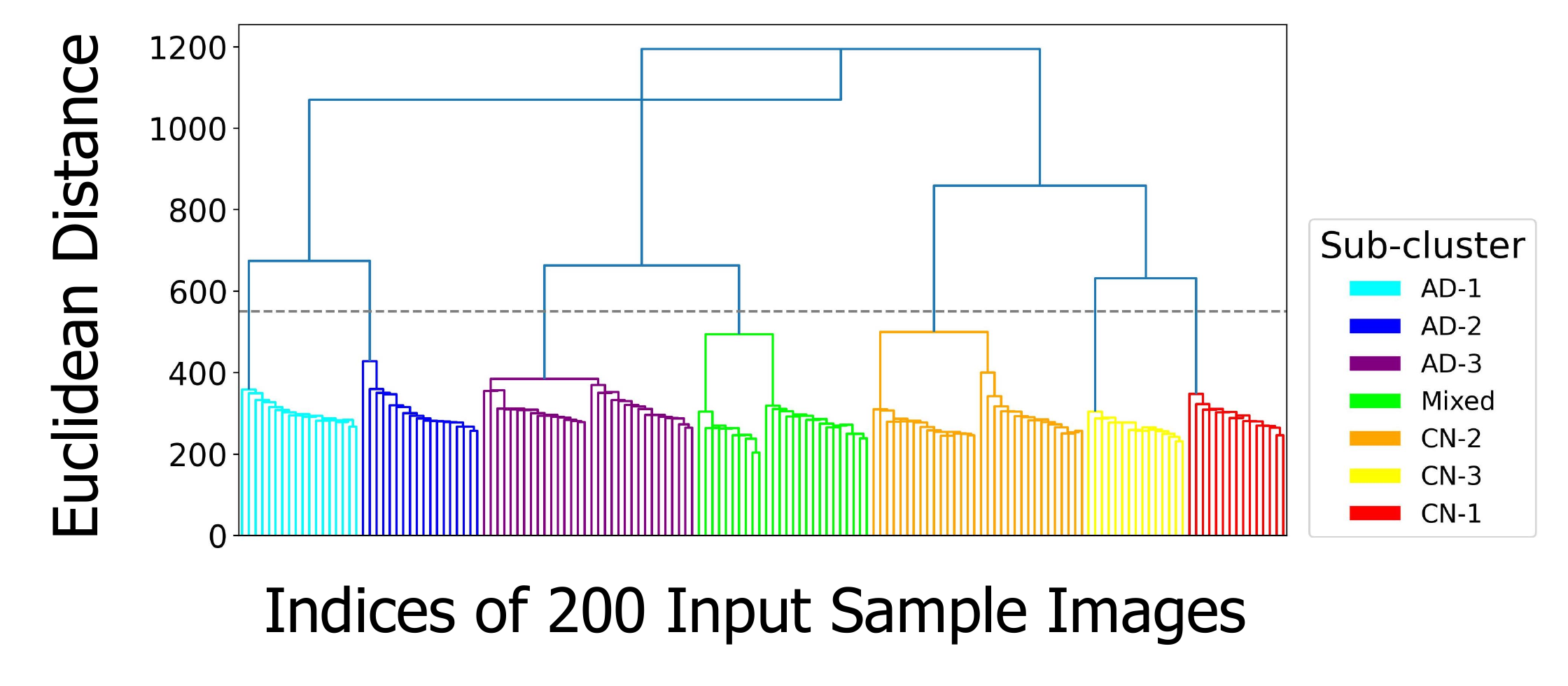}
\caption{\textbf{The results of the neuron-level hierarchical clustering.} Dendrogram of agglomerative connectivity constraint clustering with ward's linkage represents the relationships of similarity among a
clade of AD subjects.}
\label{fig: S3}
\end{figure}

\begin{figure}[!htbp]
\centering
\includegraphics[width=1\linewidth]{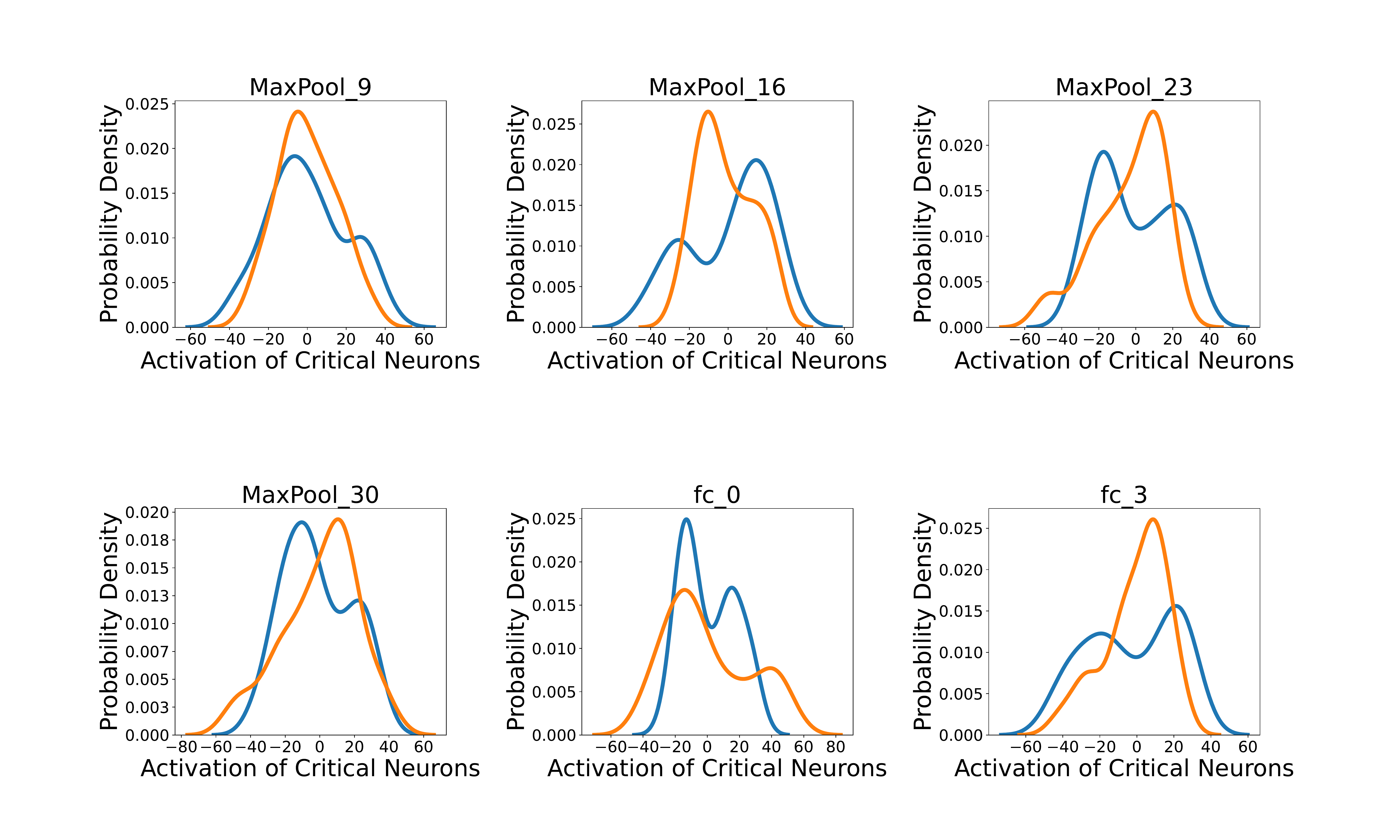}
\caption{\textbf{Exploring activation pattern of critical neurons.} Kernel density estimate (KDE) applied on top of two-dimensional embedded space (lower-dimensional representation of the critical neurons' activation at different hidden layers) obtained by t-distributed stochastic neighbor embedding (t-SNE) and unveils multiple Gaussian distributions embedded in the activation pattern of critical neurons corresponding to distinctive sub-cluster(s) associated with AD or NC target class (blue and orange curves) of the disease.}
\label{fig: S4}
\end{figure}

\begin{figure}[!htbp]
\centering
\includegraphics[width=1\linewidth]{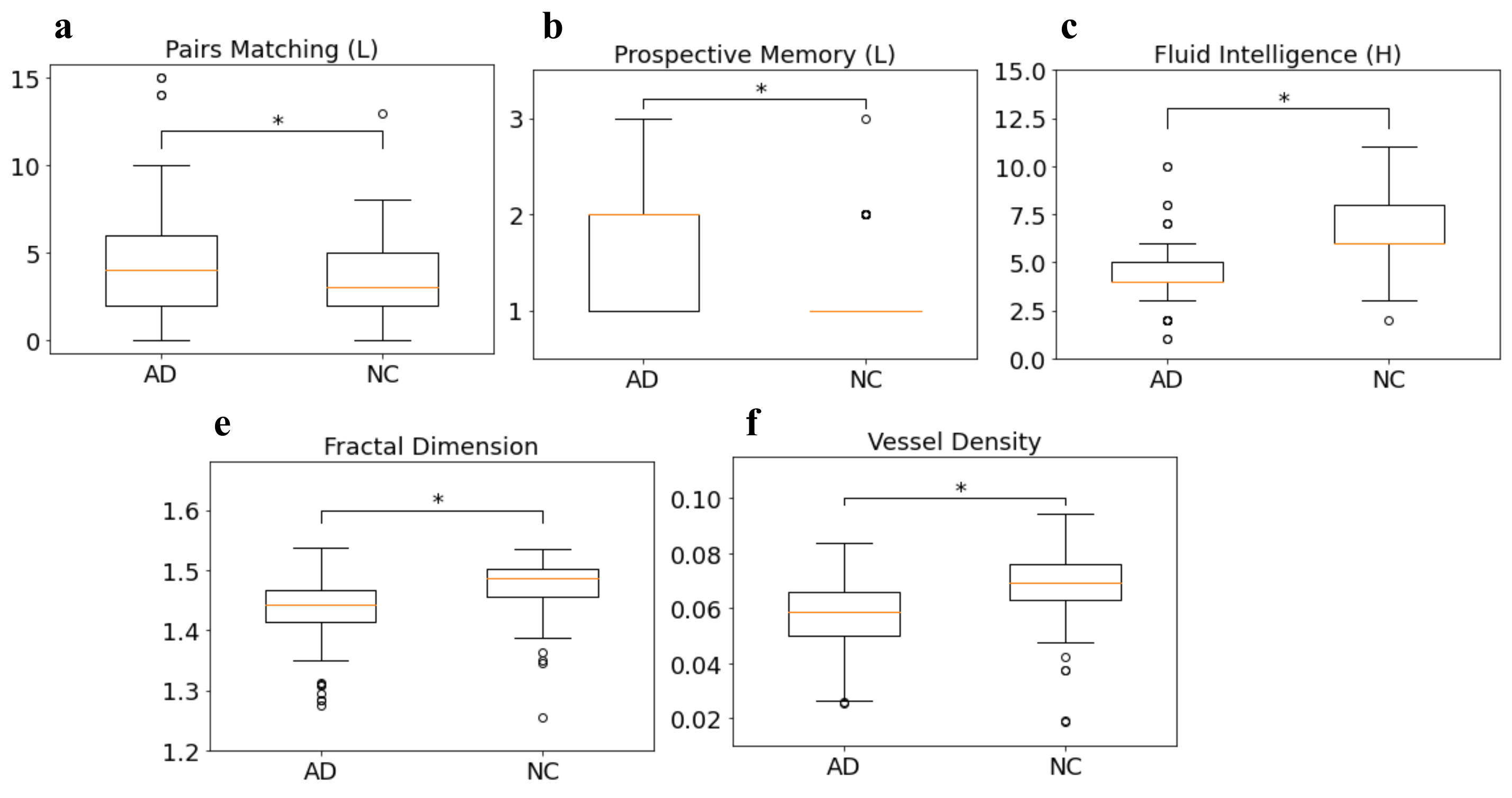}
\caption{\textbf{Box-plot comparisons between AD and NC groups of cognitive and vascular features.} (*) indicates statistical significance ($p < 0.01$) by two-tailed significance tests.}
\label{fig: S5}
\end{figure}

\newpage

\section{Supplementary Tables}

\begin{table}[!htbp]
\resizebox{\columnwidth}{!}{
\centering
\begin{tabular}{|c|c|c|c|c|c|c|c|c|}
\hline
 & CN-1 & CN-2 & CN-3 & Mixed & AD-1 & AD-2 & AD-3 & Total\\ 
\hline 
AD & 0 & 0 & 0 & 31 & 21 & 30 & 18 & 100 \\ 
\hline
CN & 15 & 39 & 15 & 31 & 0 & 0 & 0 & 100 \\
\hline
\end{tabular}}
\caption{\textbf{Results of partitioning data samples by LAVA.} In this study, the granularity level is configured to diagnose 7 subgroups of subjects corresponding to different stages of AD progression. In total 3 AD groups, 3 CN group and 1 Mixed group are identified by the LAVA framework.} 
\label{table: table1}
\end{table}

\begin{table}[!htbp]
\centering
\begin{tabular}{l  c c c } 
\hline
\textbf{Baseline Characteristics} &
\textbf{AD Group} & \textbf{HC Group} &  \textbf{pvalue} \\

\hline
N & 61 & 80 & -- \\ 
\hline
Age, mean (SD), years & 64.5 (3.6) & 63.9 (4.1) & 0.35 \\ 
\hline
Gender, No. (\%) &   & & 0.94 \\ 
\hspace{3mm} Male & 27 (44) & 34 (45) &  \\
\hspace{3mm} Female & 34 (56) & 44 (55) & \\
\hline
Ethnicity, No. (\%) & & & 0.74 \\ 
\hspace{3mm} White & 56 (91.8)  & 78 (97.5) &  \\
\hspace{3mm} Others & 5 (8.2) & 2 (2.5) &  \\
\hline
Eye Problems (\%) &  &  & 0.98 \\
\hspace{3mm} Yes & 10 (16.4) & 13 (16.3) &  \\
\hspace{3mm} No & 51 (83.6) & 67 (83.8) &  \\
\hline
Visual Acuity, mean (SD), LogMAR & 0.11 (0.20) & 0.06 (0.20) & 0.16 \\
\hline

\hline

Townsend Indices, mean (SD) & -1.43 (3.30) & -1.55 (2.61) & 0.81 \\
\hline

Diabetes-Obesity, No. (\%) &  &  & 0.19  \\ 
\hspace{3mm} Yes & 18 (29.5) & 16 (20) &  \\
\hspace{3mm} No & 43 (70.49) & 64 (80) &  \\
\hline
Smoking Status, No. (\%) &  &  & 0.13 \\

\hspace{3mm} Yes & 33 (54.1)& 33 (41.3) &  \\
\hspace{3mm} No & 28 (45.9) & 47 (58.8) &  \\
\hline
Alcohol Status, No. (\%) &  &  & 0.18  \\ 
\hspace{3mm} Yes & 60 (98.4) & 75 (93.8) &  \\
\hspace{3mm} No & 1 (1.6) & 5 (6.3)  &  \\
\hline
History of Stroke, No. (\%) &  &  & 0.01* \\ 
\hspace{3mm} Yes & 8 (13.1)  & 2 (2.5) &  \\
\hspace{3mm} No & 53 (86.9) & 78 (97.5) &  \\

\hline
\end{tabular}
\caption{\textbf{Baseline characteristics of the study populations}. P-values conducted on continuous data are computed by the Student's t-test. Categorical variables are computed by Pearson's Chi-squared test. * indicates statistically significant ($p < 0.05$). }
\label{table: study_population}
\end{table}

\end{document}